\documentclass[10pt,twocolumn,letterpaper]{article}

\usepackage[pagenumbers]{cvpr} 

\usepackage{graphicx}
\usepackage{amsmath}
\usepackage{amssymb}
\usepackage{booktabs}
\usepackage{tabularx}
\usepackage[pagebackref,breaklinks,colorlinks]{hyperref}

\usepackage[capitalize]{cleveref}
\crefname{section}{Sec.}{Secs.}
\Crefname{section}{Section}{Sections}
\Crefname{table}{Table}{Tables}
\crefname{table}{Tab.}{Tabs.}

\def\cvprPaperID{*****} 
\def\confName{CVPR}
\def\confYear{2023}

\begin{document}

\title{ZJU ReLER Submission for EPIC-KITCHEN Challenge 2023: \\
Semi-Supervised Video Object Segmentation}

\author{
  Jiahao Li, Yuanyou Xu,  Zongxin Yang,  Yi Yang, Yueting Zhuang\\
		 ReLER, CCAI, Zhejiang University\\
      {\tt\small
\{xljh,yoxu,yangzongxin,yangyics,yzhuang\}@zju.edu.cn}
      }
\maketitle

\begin{abstract}
    The Associating Objects with Transformers (AOT) framework has exhibited exceptional performance in a wide range of complex scenarios for video object segmentation~\cite{yang2021associating, yangdecoupling}.
    In this study, we introduce MSDeAOT, a variant of the AOT series that incorporates transformers at multiple feature scales. Leveraging the hierarchical Gated Propagation Module (GPM), MSDeAOT efficiently propagates object masks from previous frames to the current frame using a feature scale with a stride of 16. Additionally, we employ GPM in a more refined feature scale with a stride of 8, leading to improved accuracy in detecting and tracking small objects.
    Through the implementation of test-time augmentations and model ensemble techniques,
    we achieve the top-ranking position in the EPIC-KITCHEN VISOR Semi-supervised Video Object Segmentation Challenge.
 \end{abstract}

\section{Introduction}
\label{sec:intro}
Video object segmentation is a vital task in computer vision that involves segmenting and isolating specific objects of interest in each frame of a video sequence. This task aims to provide pixel-level masks or contours delineating the target object's boundaries in each frame.

\begin{figure*}[ht]
    \vspace{-2mm}
        \centering
        \includegraphics[width=0.8\linewidth]{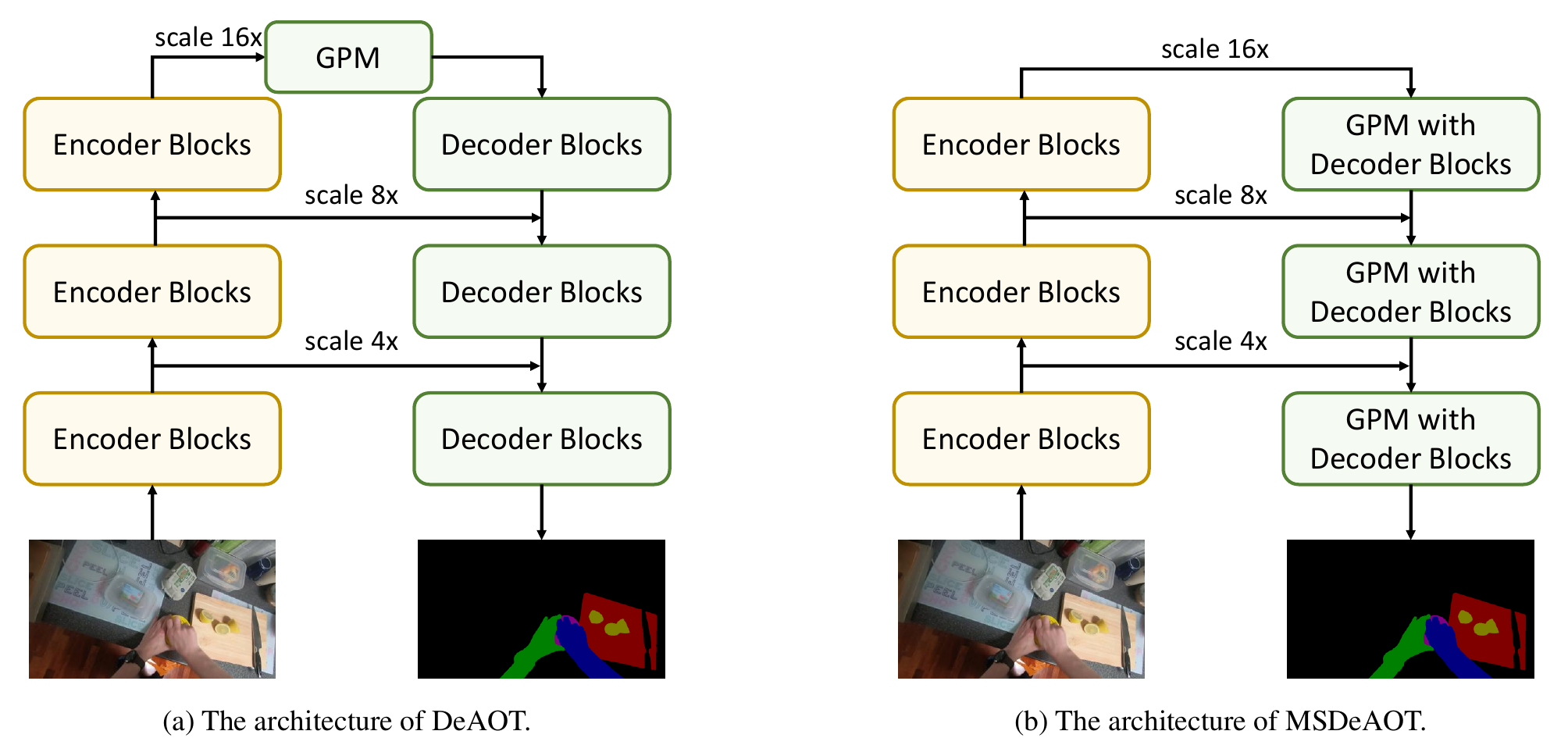}
        \caption{The above two pictures compare the architectures of DeAOT and MSDeAOT. The main difference is that the GPM module is combined with the decoder to form a multi-scale architecture in the MSDeAOT.}
        \label{fig:method_overview}
    \vspace{-2mm}
\end{figure*}

Semi-supervised Video Object Segmentation (SVOS) has garnered significant attention in recent years, with a particular emphasis on learning-based methodologies~\cite{li2023unified, cheng2023segment, zhou2022survey, li2022locality}.
In this context, a prevailing approach involves pixel matching across frames to extract crucial insights regarding the target objects.
Notably,  PLM~\cite{shin2017pixel} emerges as a pioneering matching-based SVOS method. By performing multi-scale matching between the previous and target frames, PLM successfully segments the target object. To address the challenge of ambiguous backgrounds, the method specifically focuses on the region surrounding the object in the previous frame. 
FEELVOS~\cite{voigtlaender2019feelvos} leverages both global and local matching based on pixel-wise embeddings to facilitate information transfer across frames. These matching results are further integrated with semantic features to predict the final segmentation outcome. 
CFBI(+)~\cite{yang2020collaborative,yang2021collaborative} employs the first frame and the previous frame as references, treating foreground and background regions with equal importance, implicitly enhancing the discriminability of encoded features.
These pioneering techniques contribute to the ongoing development of SVOS, offering valuable insights and potential avenues for further exploration in the field.

AOT~\cite{yang2021associating} stands as another prominent contribution in this field. Unlike previous methods which segment multiple objects one by one and merge results by post ensemble, AOT presents an innovative identification mechanism that encompasses the encoding, matching, and decoding of multiple objects. By incorporating transformers, AOT effectively associates objects across frames, fostering a deeper exploration and exploitation of inter-object relationships. 
Building upon this foundation, DeAOT~\cite{yangdecoupling}, a variant of AOT, employs a hierarchical Gated Propagation Module (GPM) to independently propagate object-agnostic and object-specific embeddings from previous frames to the current frame. This novel approach effectively preserves object-agnostic visual information within the deep transformer layers.

Although the aforementioned methods have demonstrated remarkable performance on conventional datasets such as YouTube-VOS~\cite{xu2018youtube} and DAVIS~\cite{pont20172017}, there remain several challenges that demand attention when dealing with egocentric VISOR~\cite{VISOR2022, Damen2022RESCALING}. \textbf{\romannumeral1}) Egocentric videos often exhibit rapid camera movements, posing a significant obstacle. \textbf{\romannumeral2}) In the VISOR dataset, frames are not uniformly sampled at a fixed frame rate, and the time intervals between frames can span up to 1 or 2 seconds. \textbf{\romannumeral3}) Frequent object-hand interactions in egocentric scenarios introduce issues such as occlusion and motion blur, necessitating effective handling strategies.

To address above issues, we propose MSDeAOT, another AOT-based VOS model. MSDeAOT leverages the GPM~\cite{yangdecoupling} to propagate object masks from previous frames to the current frame. Specifically, we employ GPM in 2 feature scales with strides  of 16 and 8, respectively. 
This multi-scale strategy effectively addresses the aforementioned challenges, enabling MSDeAOT to achieve $\mathcal{J} \& \mathcal{F} $ score of 89.0\% in EPIC-KITCHENS VISOR Semi-supervised Video Object Segmentation Challenge.

\section{Method}
\label{sec:method}

In this section, we present our main method in detail. We begin by providing a brief overview of the DeAOT model to familiarize the reader with the foundational concepts. Subsequently, we delve into the architecture design of our novel MSDeAOT model, outlining its key features and advancements. By elucidating these aspects, we aim to offer a comprehensive understanding of our approach and its contributions.

\subsection{Revisiting DeAOT}
\label{sec:deAOT}

DeAOT~\cite{yangdecoupling}, an adaptation of AOT, preserves the fundamental IDentification mechanism of AOT~\cite{yang2021associating}. Consequently, DeAOT exhibits the ability to effectively handle multiple objects concurrently. Diverging from AOT, which consolidates the visual (object-agnostic) and ID (object-specific) embeddings within a shared embedding space, DeAOT adopts a decoupled approach. This entails employing separate propagation processes for each embedding type while maintaining shared attention maps. 
By decoupling these embeddings, DeAOT achieves enhanced flexibility and discriminative power in handling object representations.

The Gated Propagation Module (GPM) plays a pivotal role within the DeAOT framework. In contrast to the LSTT block utilized by AOT, which leverages multi-head attention for propagation, GPM employs a Gated Propagation Function to seamlessly fuse and propagate both object-agnostic and object-specific embeddings. Additionally, GPM employs single-head attention to match objects and effectively propagate information across frames. By incorporating these distinctive mechanisms, GPM enhances the overall performance of the DeAOT model, enabling more accurate and efficient object segmentation.

\subsection{Multi-Scale DeAOT}

The whole architecture of MSAOT as shown in~\cref{fig:method_overview}\textcolor{red}{b} follows an encoder-decoder design similar to classical segmentation networks like U-Net. The encoder consists of multiple blocks that down-sample the input feature maps, yielding features at different scales. These encoder blocks provide multi-scale features that are crucial for accurate object tracking and segmentation.

In the decoder, unlike the FPN module employed in DeAOT (~\cref{fig:method_overview}\textcolor{red}{a}), the Gated Propagation Module (GPM) is integrated with multiple decoder blocks to establish the multi-scale stages of MSDeAOT. Each scale's feature maps from the encoder are fed into the corresponding stage, where the GPM module takes charge of matching the current frame with memory frames and aggregating mask information from the memory frames. The decoder blocks then decode this information.

This innovative design of multi-scale stages brings notable benefits. It effectively harnesses the potential of feature maps at different scales, in contrast to the FPN module used in DeAOT, where multi-scale feature maps solely serve as shortcut connections for residual structures. Specifically, in DeAOT, only the feature maps at the smallest scale are utilized for matching across memory frames using the GPM module. In contrast, MSDeAOT comprehensively engages feature maps from multiple scales during the matching process, thereby enhancing performance and enabling finer details of objects to be captured.
\section{Implementation Details}


In MSDeAOT, we employ ResNet-50~\cite{he2016deep} and Swin Transformer-Base~\cite{liu2021swin} as the backbones for the encoder. While ResNet-50 offers a lightweight option, Swin Transformer-Base achieves superior performance.  For the decoder, the MSDeAOT model incorporates GPM modules in multiple stages. Specifically, we set the number of layers in the GPM to 2 for the 16$\times$ scale stage and 1 for the 8$\times$ scale stage. To save computational resources, we exclude the 4$\times$ scale feature maps and instead duplicate the 16$\times$ scale feature maps twice to form the feature pyramid.

The training process comprises two phases, following the AOT framework. In the initial phase, we pre-train the model using synthetic video sequences generated from static image datasets~\cite{cheng2014global,everingham2010pascal,hariharan2011semantic,lin2014microsoft,shi2015hierarchical} by randomly applying multiple image augmentations~\cite{oh2018fast}.  In the subsequent phase, we train the model on the train and val sets of the VISOR dataset~\cite{VISOR2022}, incorporating random video augmentations~\cite{yang2020collaborative}.

During MSDeAOT training, we employ 8 Tesla V100 GPUs with a batch size of 16.  For pre-training, we use an initial learning rate of $4 \times 10^{-4}$
for 100,000 steps. For main
training, the initial learning rate is set to $2 \times 10^{-4}$, and the
training steps are 100,000.
The learning rate gradually decays to $1 \times 10^{-5}$ using a polynomial decay schedule~\cite{yang2020collaborative}.
\section{EPIC-Kitchens Challenge: Semi-Supervised Video Object Segmentation}
\label{sec:challenge}
\subsection{Ablation study on VISOR val set}

\begin{table}[t]
      \centering
      \small
      \setlength{\tabcolsep}{1pt} %
        \begin{tabularx}{\linewidth}{>{\raggedright\arraybackslash}p{3cm} >{\centering\arraybackslash}X >{\centering\arraybackslash}X >{\centering\arraybackslash}X }
     
    \hline
    Method & $\mathcal{J} \& \mathcal{F} $ & $\mathcal{J}$ & $\mathcal{F}$  \\
    \hline
    R50-MSDeAOT & 84.0 & 81.7  & 86.3  \\
    SwinB-MSDeAOT & 85.6 & 83.2  & 87.9  \\
    \hline

      \end{tabularx}
      \caption{Comparisons on VISOR val set. Models are trained on VISOR train set.}
      \label{table:tab1_val}
\end{table}

\begin{table}[t]
    \centering
    \small
    \setlength{\tabcolsep}{1pt} %
      \begin{tabularx}{\linewidth}{>{\raggedright\arraybackslash}p{2.2cm} >{\centering\arraybackslash}p{5cm} >{\centering\arraybackslash}X}
   
  \hline
  Method & Training data & $\mathcal{J} \& \mathcal{F} $   \\
  \hline
  R50-DeAOTL & VISOR~\cite{VISOR2022} & 83.1   \\
  R50-DeAOTL & VISOR~\cite{VISOR2022},YouTube-VOS~\cite{xu2018youtube} & 82.9   \\
  R50-DeAOTL & VISOR~\cite{VISOR2022},DENSE VISOR~\cite{VISOR2022} & 82.5   \\
  R50-MSDeAOT&  VISOR~\cite{VISOR2022}  & 84.0   \\
  R50-MSDeAOT&  VISOR~\cite{VISOR2022},VOST~\cite{tokmakov2023breaking}  & 83.7   \\
  \hline

    \end{tabularx}
    \caption{Ablation study of training data. Models are evaluated on VISOR val set.}
    \label{table:tab2_dataset}
\end{table}

\begin{table}[t]
      \centering
      \small
      \setlength{\tabcolsep}{1pt} %
        \begin{tabularx}{\linewidth}{>{\raggedright\arraybackslash}p{3cm} >{\centering\arraybackslash}X >{\centering\arraybackslash}X >{\centering\arraybackslash}X }
     
    \hline
    Ensemble Method & $\mathcal{J} \& \mathcal{F} $ & $\mathcal{J}$ & $\mathcal{F}$  \\
    \hline
    Logits Meaning & 88.9 & 86.7  & 91.0  \\
    Masks Voting & 89.0 & 86.9  & 91.2  \\
    \hline

      \end{tabularx}
      \caption{Results on VISOR test set. Models are trained on VISOR train and val set.}
      \label{table:tab2_test}
\end{table}

We train our MSDeAOT on two backbones, ResNet-50~\cite{he2016deep} and Swin Transformer-Base~\cite{liu2021swin}, and evaluate them on the VISOR val set. As shown in~\cref{table:tab1_val}, SwinB-MSDeAOT achieves better performance than R50-MSDeAOT, with a $\mathcal{J} \& \mathcal{F}$ score of 85.6. 

As for training data, we train R50-DeAOTL and R50-MSDeAOT on multiple datasets and evaluate them on VISOR val set. As shown in~\cref{table:tab2_dataset}, R50-MSDeAOT trained on VISOR achieves the best performance, with a $\mathcal{J} \& \mathcal{F}$ score of 84.0 and more training data has no improvement.

\subsection{Model Ensemble}

To achieve the best performance on hidden test, we train 3 models, \ie, SwinB-DeAOTL~\cite{yangdecoupling}, R50-MSDeAOT, and SwinB-MSDeAOT on VISOR train and val set. We then ensemble 
the predictions of these 3 models to obtain the final results. 
As for test-time
augmentations, both multi-scale test and flip test are used.
The scales are {1.2$\times$, 1.3$\times$, 1.4$\times$} and each scale includes
non-flipped and flipped test.

\noindent\textbf{Logits Meaning.} 
During the inference process of each model, we refrain from storing the masks themselves. Instead, we retain the logits, which represent the probabilities of each pixel belonging to a specific object. Once inference is completed across multiple models, we acquire multiple sets of logits. These sets are subsequently weighted, averaged, and evaluated. The final label is determined by selecting the set with the highest probability value, thus consolidating the collective predictions from the ensemble of models.
However, saving logits requires a large disk space, we can only apply two logtis for ensemble.

\noindent\textbf{Mask Voting.} 
First, we adopt a multi-model approach to perform inference, wherein each model produces a mask for an image. As a result, each pixel is associated with multiple labels. To determine the ultimate label for each pixel, we employ a weighted voting scheme that aggregates the labels. This process ensures the derivation of a final label that captures the collective information from the multiple models.
As saving masks requires much less disk space than saving logits, we can apply more results of different models for ensemble.

\section{Conclusion}

In this paper, we propose MSDeAOT, a variant of AOT, for semi-supervised video object segmentation. MSDeAOT leverages the hierarchical Gated Propagation Module (GPM) to independently propagate object-agnostic and object-specific embeddings from previous frames to the current frame. 
MSDeAOT shows remarkable performance on the EPIC-KITCHENS VISOR Semi-supervised Video Object Segmentation Challenge with $\mathcal{J} \& \mathcal{F} $ of 89.0\% on the test set.

{\small
\bibliographystyle{ieee_fullname}
\bibliography{main}
}

\end{document}